\documentclass[letterpaper,10pt,conference]{ieeeconf}
\IEEEoverridecommandlockouts

\usepackage[dvipsnames]{xcolor}
\usepackage{amsmath,amsfonts}
\usepackage[framemethod=tikz]{mdframed}
\usepackage{hyperref}
\hypersetup{hidelinks}
\usepackage{cleveref,tabularx}
\Crefname{figure}{Fig.}{Figs.}
\Crefname{section}{Sec.}{Secs.}
\Crefname{table}{Table}{Tables}
\Crefname{equation}{Eqn.}{Eqns.}

\usepackage{diagbox}
\usepackage{multirow}
\usepackage{comment}
\usepackage{tikz}
\usepackage{pgfplots}
\usepackage[normalem]{ulem}

\definecolor{forestgreen}{RGB}{34 139 34}
\definecolor{darkorchid}{RGB}{153 50 204}
\definecolor{darkorange}{RGB}{255 140 0}

\graphicspath{{figures/}{Figures/}} %Setting the graphicspath

\usepackage[per-mode=symbol]{siunitx}            % SI unit x
\newcommand{\fh}{\mbox{$\mathrm{dF}_\mathrm{hTO}$}}
\newcommand{\fhlong}{\mbox{$\Delta ^{s} F_{h,l} [\%]$}}

\usepackage{caption}
\pgfplotsset{compat=1.17}
\title{\LARGE \bf Multi-segmented Adaptive Feet for \\ Versatile Legged Locomotion in Natural Terrain}

\author{Abhishek Chatterjee\textsuperscript{1}, An Mo\textsuperscript{1}, Bernadett Kiss\textsuperscript{1}, Emre Cemal G\"onen\textsuperscript{1}, Alexander Badri-Spr\"owitz\textsuperscript{1,2}\\
\textsuperscript{1}{Dynamic Locomotion Group, Max Planck Institute for Intelligent Systems, Stuttgart, Germany} \\
\textsuperscript{2}{Department of Mechanical Engineering, KU Leuven, Leuven, 3001, Belgium}\\
    {\tt [chatterjee;mo;kiss;gonen;sprowitz]@is.mpg.de}}

\begin{document}
\maketitle
%\thispagestyle{empty}
% \pagestyle{empty} % empty % change for submission back to empty, plain to show the page numbers at the center

%%%%%%%%%%%%%%%%%%%%%%%%%%%%%%%%%%%%%%%%%%%%%%%%%%%%%%%%%%%%%%%%%%%%%%%%%%%%%%%%
\begin{abstract}
Most legged robots are built with leg structures from serially mounted links and actuators and are controlled through complex controllers and sensor feedback. In comparison, animals developed multi-segment legs, mechanical coupling between joints, and multi-segmented feet. They run agile over all terrains, arguably with simpler locomotion control. Here we focus on developing foot mechanisms that resist slipping and sinking also in natural terrain. We present first results of multi-segment feet mounted to a bird-inspired robot leg with multi-joint mechanical tendon coupling. Our one- and two-segment, mechanically adaptive feet show increased viable horizontal forces on multiple soft and hard substrates before starting to slip. We also observe that segmented feet reduce sinking on soft substrates compared to ball-feet and cylinder-feet. We report how multi-segmented feet provide a large range of viable centre of pressure points well suited for bipedal robots, but also for quadruped robots on slopes and natural terrain. Our results also offer a functional understanding of segmented feet in animals like ratite birds.
\end{abstract}
%%%%%%%%%%%%%%%%%%%%%%%%%%%%%%%%%%%%%%%%%%%%%%%%%%%%%%%%%%%%%%%%%%%%%%%%%%%%%%%%
\section{Introduction}
We envision that robust and reliable legged robots will support us not only at home and in our indoor work spaces. Ideally, they will work wherever humans go and beyond. Fields, forests, and mountains are examples of natural environments to which vehicles have limited access. Where animals roam naturally, most wheeled and tracked vehicles are hindered by terrain roughness, slopes, available space, and natural substrates. 
Towards the goal of legged robots in natural environments, here we aim to design and characterize multi-segmented robot feet mounted to a bird-inspired robot leg. With animals as models in mind, we want to equip our robots with feet that mechanically adapt their shape to comply with rigid, uneven terrain and grab into soft terrain. Bipedal robots require feet with a range of viable center of pressure (CoP) points, or they must constantly and actively be balanced \cite{ma_bipedal_2017}. This criterion disqualifies commonly mounted ball- and cylinder-shaped feet that otherwise perform well on quadruped legged robots \cite{hutter_toward_2014,grimmingerOpenTorqueControlledModular2020}.

No robots so far run as efficiently, dynamically and autonomously in natural environments as animals, which indicates a hard challenge that involves complex interactions between robot mechanics, ground mechanics, and robot control and sensing \cite{yang_grand_2018}. Unlike engineered streets and floors, natural terrain is non-uniform and unstructured. Soft substrates react compliantly, while sand and loose pebbles act like a viscous fluid \cite{liSensitiveDependenceMotion2009}. Rubber-coated footwear that provides high friction indoors suddenly turns slippery on wet surfaces. Rigid but uneven substrates like rocks and roots are challenging as one must seek secure footholds \cite{lee_learning_2020}. Feet will sink into sand and snow until they are immersed and need to be pulled out, potentially destabilizing locomotion patterns \cite{hubicki_atrias}. 
%%%%%%%%%%%%%%%%%%%%%%%%%%%%%%%%%%%%%%%%%%%%%%%%%%%%%%%%%%%%%%%%%%%
\begin{figure} \centering
\includegraphics[width=0.6\columnwidth]{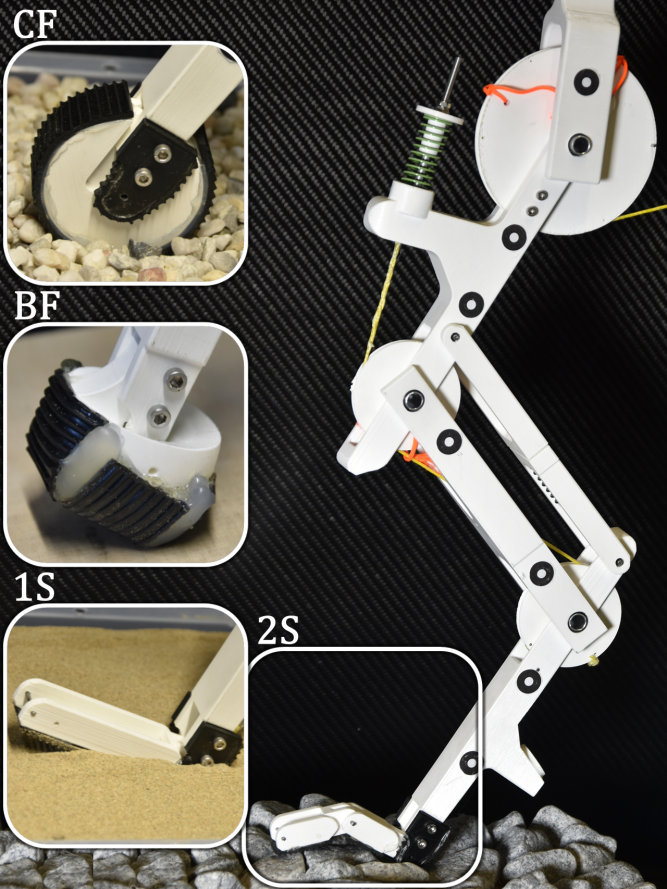}
\caption{Multi-segmented robot leg and four foot designs. Toes are parts of feet, and we will use both terms for our digitigrade robot leg. The robot leg is an enlarged copy of the BirdBot robot leg \cite{badri2022}. The leg features a global spring tendon that provides body weight support in stance phase, and exerts toe joint torques. The 2-segmented (2S) toe stands on a rough stone surface. A cylinder foot (CF) on round pebbles, a ball foot (BF) on flat wood, and a 1-segmented foot (1S) on loose sand were used in our experiments.}
\label{fig:collage}
\end{figure}

The complexity of natural substrates raises the question of how to create and maintain reliable foot-substrate contact without sliding and how to withstand high horizontal reaction forces. The main options are; a) through sensing, control, and actuation, b) through a mechanically adaptive foot, and c) by combining the above. Most controllers for legged locomotion are designed in the laboratory, i.e., optimized for flat and rigid grounds \cite{grimmingerOpenTorqueControlledModular2020,li_systematic_2010}. Controllers were developed for soft and uneven rough terrain \cite{hubicki_atrias,bosworth_robot_2016,hubicki_tractable_2016}, but require precise and high-bandwidth sensing, a good knowledge of ground conditions, and high computational effort \cite{zapolsky_inverse_2013,wu_integrated_2016,buchli_compliant_2009,ruppert_foottile_2020,grimmingerOpenTorqueControlledModular2020}. Alternatively, mechanically adaptive legs and feet have been reported to increase their legged robot's performance on slopes \cite{sprowitz_oncilla_2018}, at step-downs \cite{sprowitz_towards_2013}, and rough terrain through form adaptation, without or with only low computational costs or required sensing \cite{hauser_friction_2016,kolvenbach_traversing_2021,catalanoAdaptiveFeetQuadrupedal2022}. 

Animals pass over natural terrains seemingly effortless \cite{daley_biomechanics_2008,spence_insects_2010}. Adaptive feet mechanisms are known for lizards running on water \cite{hsieh_three} and land \cite{liMultifunctionalFootUse2012}. Legs and feet of horse-like animals feature mechanisms that increase the friction between hoofs and substrates \cite{oosterlinckPreliminaryStudyPressure2014,abad_role_2016}. Flat feet or flat-footed toe structures (plantigrade and digitigrade lower limbs, \cite{carrano_morphological_1997}) are required to supply a viable centre of pressure (CoP) in bipeds, but not necessarily in quadruped animals, as hoofed (unguligrade) animals show. Yet in environments with soft substrates (snow, sand, mud), we also observe that quadruped animals developed large foot pads, likely to counter sinking~\cite{nyakatura_reverse_2019}, support accelerations, and climbing slopes of soft substrate. Well-known examples are the feet of camels and snow leopards.
The specialized ankle joint and foot arch of humans supports locomotion robustness by storing energy \cite{kiss_gastrocnemius_2022,buchmann_power_2022,bruening_midtarsal_2018}, and by adapting intra-foot joint torques \cite{liu_higher_2016,chen_bevel2021}. 
Birds can stand, potentially fully passive, thanks to multi-joint leg-toe muscle-tendon coupling \cite{chang_mechanical_2017,badri2022}. 
Especially smaller birds utilize a tendon-based mechanical coupling between leg- and toe-joints to passively perch \cite{watson_mechanism_1869,doyle_avian,nadan_bird,backus_mechanical_2015}. 
Therefore, instead of relying on the closed-loop action of a sensor-motor apparatus, toe grasping forces can be produced by a tendon-based mechanism. 
As a result, grasping and attachment actions are robustly executed in a few ten milliseconds~\cite{lee_self_engaging_2018,roderick_bird_inspired_2021}. 
Among model animals, large bipedal birds stand out as they effectively manipulate soft surfaces \cite{turnerItLoopShared2020}. Their tendon-loaded, elongated phalanges (toe segments) allow them to shift their CoP, balance in several directions, adapt their toe shape to uneven terrain and claw into soft substrates \cite{zhang_finite_2013,zhang_finite_2016}. 

\begin{figure}[t] % figure
\includegraphics[width=0.9\columnwidth]{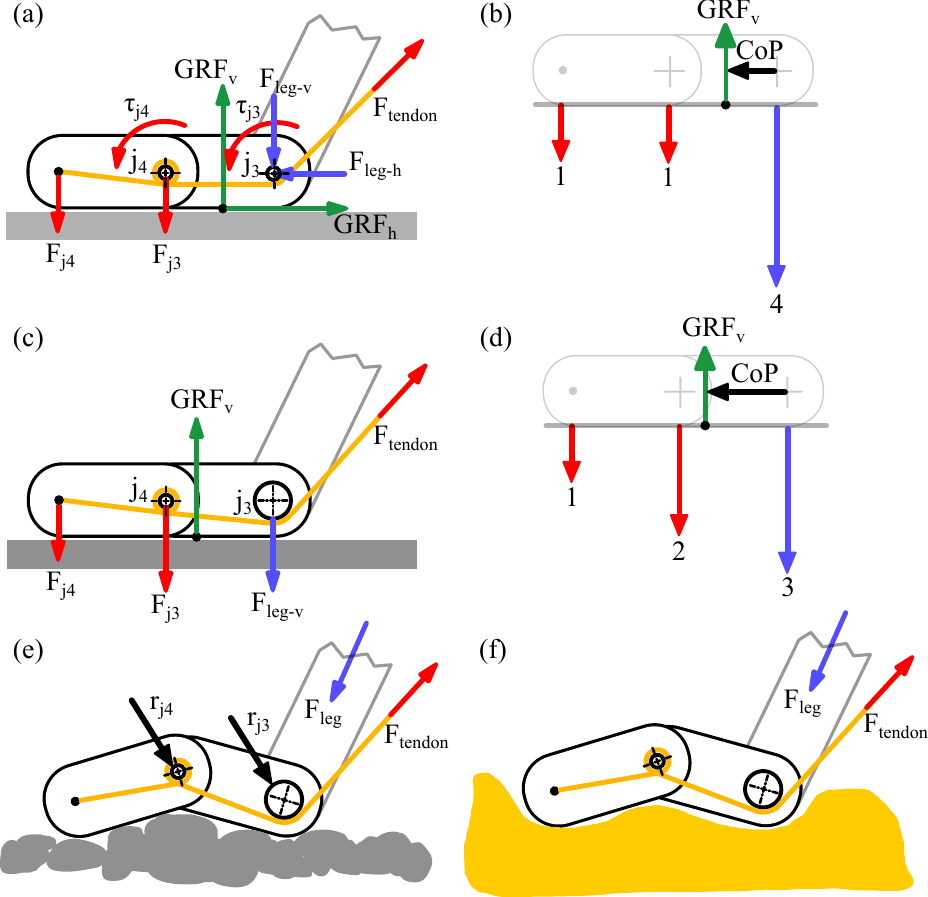}
\caption{Force and shape interaction of a 2-segmented foot on (a, c) flat wood, (e) uneven stone and (f) sand. The tendon force $F_{tendon}$ leads to joint torques $\tau_{j3}$ and  $\tau_{j4}$ through toe pulleys $r_{j3}$ and $r_{j4}$, respectively. Joint torques push toe segments against the ground with forces $F_{j3}$, $F_{j4}$. A larger toe pulley (c vs. a) $r_{j3}$ increases $F_{j3}$ and shifts the CoP towards the toe-tip (d vs. b). }
\label{fig:SegFootCoupleExample}
\end{figure}

In the following, we utilize two leg designs. First, a previously published, bird-inspired robot leg that mechanically couples leg length forces with toe-joint torques (LL, `leg-length' coupling configuration, \Cref{fig:collage}). As 1-segment and 2-segment toes, we test adaptive feet on hard and soft substrates. We also test ball feet (BF) and cylinder feet for comparison. With a second leg design (TJ, `toe joint' coupling configuration, \cite{kurokawa_introduction_2018,kiss_gastrocnemius_2022}), we reduce in-stance toe-joint torques by decoupling leg-length forces from toe torques. Instead, the TJ leg inserts toe-joint torques only by leg {\em vaulting}--leg angle motion during locomotion. Our ideal foot should be \emph{adaptive} and
a) conform to uneven hard surfaces. This shape change will help
b) increase viable horizontal ground reaction forces for high fore-aft acceleration \cite{williams_pitch_2009}, and the ability to climb rough terrain and slopes \cite{buchli_compliant_2009}. Adaptive feet should 
c) grab into substrates to resist horizontal sliding when possible and 
d) resist sinking into soft substrates. 
Equally, robot feet should be e) \emph{versatile}. Instead of a foot that performs well on a single substrate, we prefer one that performs well on various soft and hard substrates. \Cref{fig:SegFootCoupleExample} shows a 2-segmented foot adapting to flat-rigid (wood), uneven-rigid (stone), and soft and granular surfaces like sand. 

In this work, we provide for the first time a systematic test and comparison between multi-segment feet (1- and 2-segment), ball-feet, and cylinder-feet (\Cref{{fig:feetcad}}) on soft and hard substrates. We identify and measure the onset of sliding in form of peak viable horizontal forces, and we report how much individual legs resist sinking into soft substrates. We mount combination of two leg designs and four foot types into our setup and push legs over a force plate covered with four substrates types.
% The manuscript is organized as follows: \Cref{sec:methods} presents leg design details, the method for selection of parameters, and the experimental setup. \Cref{sec:results} reports results obtained from experiments. In \Cref{sec:discussion,sec:conclusion} we interpret and summarize our work.
%%%%%%%%%%%%%%%%%%%%%%%%%%%%%%%%%%%%%%%%%%%%%%%%%%%%%%%%%%%%%%%%%%%%%%%%%%%%%%%%
%%%%%%%%%%%%%%%%%%%%%%%%%%%%%%%%%%%%%%%%%%%%%%%%%%%%%%%%%%%%%%%%%%%%%%%%%%%%%%%%
%%%%%%%%%%%%%%%%%%%%%%%%%%%%%%%%%%%%%%%%%%%%%%%%%%%%%%%%%%%%%%%%%%%%%%%%%%%%%%%%
\section{Methods}\label{sec:methods}
%%%%%%%%%%%%%%%%%%%%%%%%%%%%%%%%%%%%%%%%%%%%%%%%%%%%%%%%%%%%%%%%%%%%%%%%%%%%%%%%
\begin{figure} \centering % figure
\includegraphics[width=0.6\columnwidth]{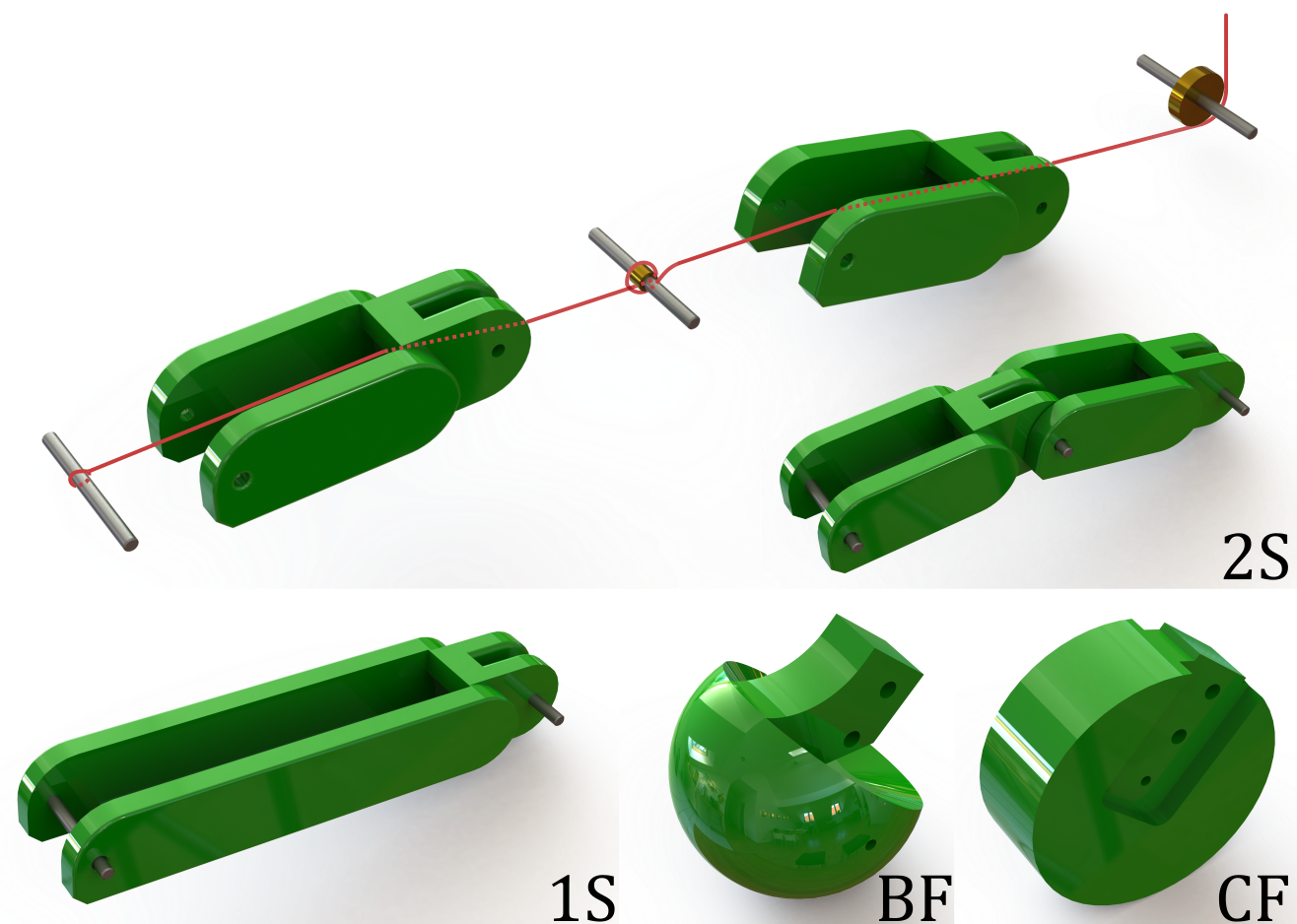}
\caption{Computer-aided-design view of a 2-segment (2S) foot, a 1-segment (1S) foot, a ball foot (BF) and a cylinder foot (CF). The red tendon applies torque to the toe joint 1 via joint j3-pulley and to the toe joint 2 via joint j4-pulley. Both toe cams can rotate freely.}
\label{fig:feetcad}
\end{figure}
\begin{figure*}[ht] 
\centering % figure
\includegraphics[width=.75\textwidth]{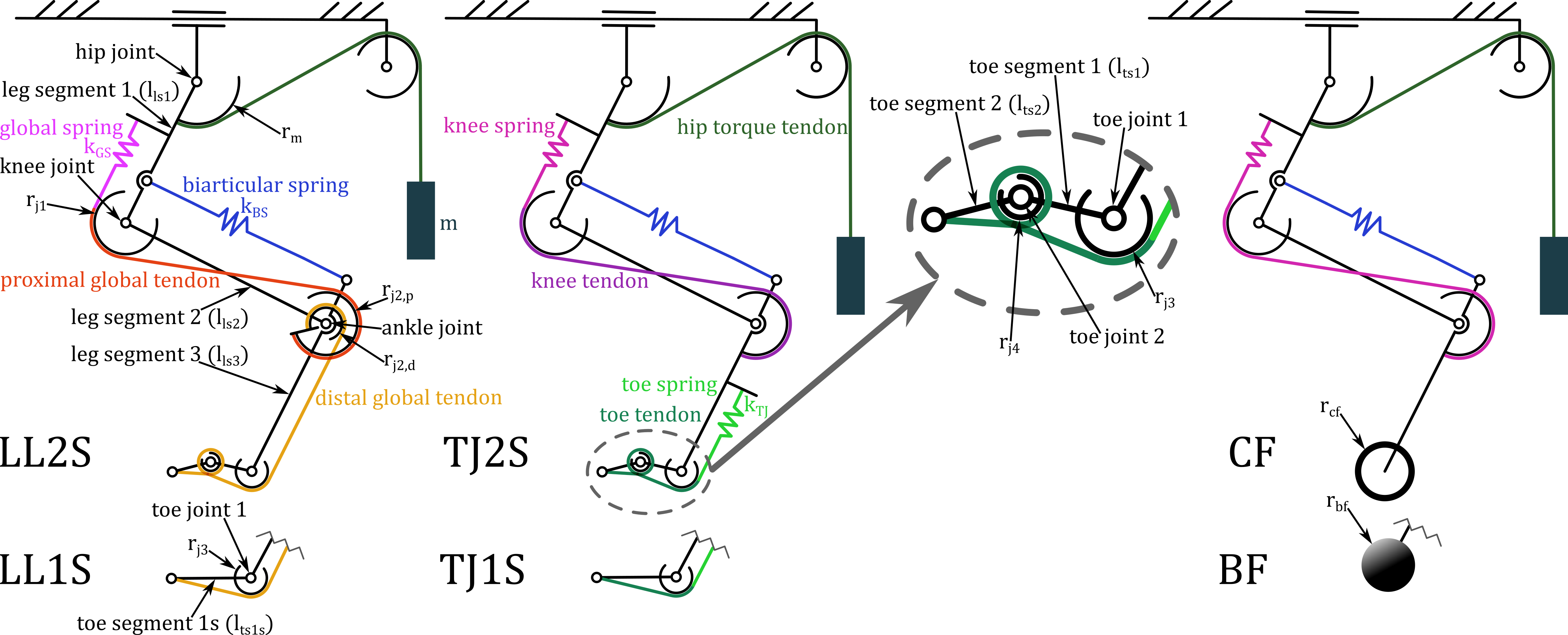}
\caption{Schematics of two leg and four segmented and non-segmented feet designs. Abbreviations (LL2S, LL1S, etc.) are explained in detail in \Cref{tab:leg_configurations}. In LL legs, leg forces are coupled mechanically with toe joint torques through the in-series network of global spring, proximal global tendon, and distal global tendon. LL legs are based on the BirdBot robot leg mechanisms \cite{badri2022}. In TJ legs, the knee spring exerts forces only into knee and ankle torque. Toe joint torque in TJ legs is decoupled from leg length changes. Instead, a toe joint spring and tendon (green) provide toe torques through leg angle motion (leg vaulting). Cylinder- and ball-feet are mounted to the same leg configuration as TJ legs, leaving out the toe spring and tendon. 
}
\label{fig:schematic}
\end{figure*}
%%%%%%%%%%%%%%%%%%%%%%%%%%%%%%%%%%%%%%%%%%%%%%%%%%%%%%%%%%%%%%%%%%%%%%%%%%%%%%%%
\textbf{Robot design}
\label{subsec:design_of_the_experiment}
We developed a bio-inspired robot leg based on BirdBot robot \cite{badri2022} to characterize four foot designs. We implemented two tendon coupling mechanisms and tested all configurations on four substrates (\Cref{fig:collage}). A leg structure includes a 3-segment pantograph with a biarticular spring connecting segments 1 and 3 (\Cref{tab:leg_parameters}, \Cref{fig:schematic}). By varying tendon couplings, foot designs, and toe pulley radii, we tested overall nine configurations (\Cref{tab:leg_configurations}):
\begin{itemize}
    \item LL2SC1/2/3: LL-coupling with 2 toe segments and pulley configuration 1,2, or 3.
    \item LL1S: LL-coupling with 1 toe segment with 1 pulley.
    \item TJ2SC1/2: TJ-coupling with 2 toe segments, pulley configurations 1, or 2.
    \item TJ1S: TJ-coupling with 1 toe segment with 1 pulley.
    \item BF: Ball foot design.
    \item CF: Cylinder foot design.
\end{itemize}
\begin{table}[b]
\caption{Leg-toe configurations.}
\label{tab:leg_configurations}
\centering
\setlength{\tabcolsep}{5pt}
\begin{tabular}{l|cccccc}
Abbrev.  & \begin{tabular}[c]{@{}c@{}}Coupling \\ type\end{tabular}  & \begin{tabular}[c]{@{}c@{}}\# Foot/toe \\ segments\end{tabular} & \begin{tabular}[c]{@{}c@{}}$r_{j2, d}$ \\ {[}mm{]}\end{tabular} & \begin{tabular}[c]{@{}c@{}}$r_{j3}$ \\ {[}mm{]}\end{tabular} & \begin{tabular}[c]{@{}c@{}}$r_{j4}$ \\ {[}mm{]}\end{tabular} & \begin{tabular}[c]{@{}c@{}}$k_{TJ}$ \\ {[}N/mm{]}\end{tabular} \\ \hline
LL2SC1  & Leg-length  & 2                                                           & 13.5                                                            & 3.2                                                          & 1.0                                                            & -                                                              \\
LL2SC2  & Leg-length  & 2                                                           & 13.5                                                            & 3.2                                                          & 2.5                                                          & -                                                              \\
LL2SC3  & Leg-length  & 2                                                           & 10.2                                                            & 9.0                                                            & 1.0                                                            & -                                                              \\
LL1S    & Leg-length  & 1                                                           & 13.5                                                            & 3.2                                                          & -                                                            & -                                                              \\
TJ2SC1  & Toe joint & 2                                                           & 13.5                                                            & 3.2                                                          & 1.0                                                            & 4.0                                                           \\
TJ2SC2  & Toe joint & 2                                                           & 13.5                                                            & 3.2                                                          & 2.5                                                          & 4.0                                                           \\
TJ1S    & Toe joint & 1                                                           & 13.5                                                            & 3.2                                                          & -                                                            & 4.0                                                           \\
BF      & -           & 1                                                           & 13.5                                                            & -                                                            & -                                                            & -                                                              \\
CF      & -           & 1                                                           & 13.5                                                            & -                                                            & -                                                            & -                                                             
\end{tabular}
\end{table}

%%%%%%%%%%%%%%%%%%%%%%%%%%%%%%%%%%%%%%%%%%%%%%%%%%%%%%%%%%%%%%%%%%%%%%%%%%%%%%%%
%%%%%%%%%%%%%%%%%%%%%%%%%%%%%%%%%%%%%%%%%%%%%%%%%%%%%%%%%%%%%%%%%%%%%%%%%%%%%%%%
\begin{figure}
    \centering
    \includegraphics[scale=0.9]{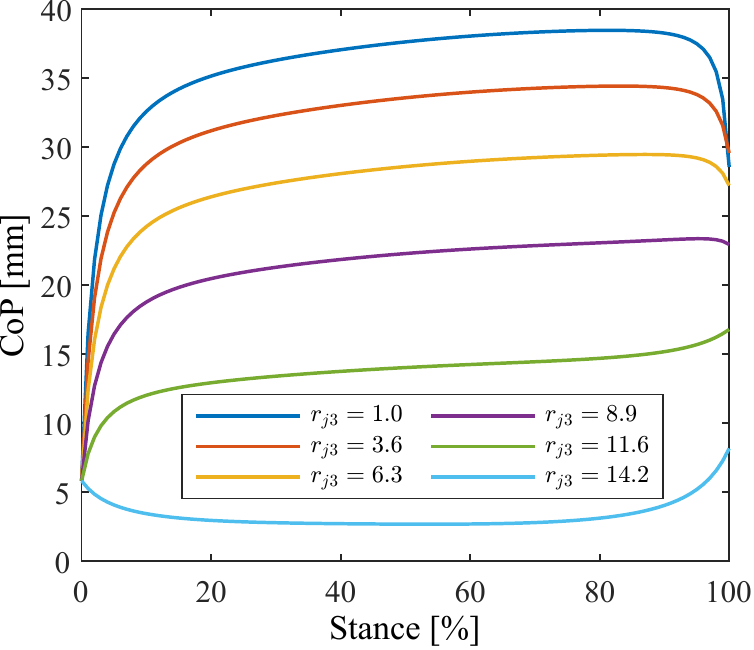}
    \caption{Center of pressure (CoP) simulation on flat ground with varying foot-pulley radius $r_{j3}$. The simulated leg pole vaults over a 1-segment foot on a horizontal hip trajectory like in hardware experiments. The foot reference point for the CoP is shown in \Cref{fig:SegFootCoupleExample}b.}
    \label{fig:CoPSim}
\end{figure}
In the LL configuration (\Cref{fig:schematic}), a tendon connects the tip of the last (distal) toe segment to the global spring tendon through multiple joint pulleys. Hence, shortening the leg length compresses the global spring. In turn, the global tendon is carrying forces, which are inserted as torque into toe joints according to joint pulley radii ($\tau=F_\mathrm{tendon} \cdot r_\mathrm{pulley}$, \cite{biewener_scaling_1989}). Note that the two pulleys ($r_{j2,p}, r_{j2,d}$) of the ankle joint are rigidly connected, as documented for the BirdBot robot. In the TJ configuration, a tendon connects the toe-tip to the toe-spring mounted to the leg segment 3. Toe joint torques then result from leg vaulting over the toe joint, which increases toe-angle. Instead of the global spring tendon, the knee spring tendon connects leg segment 1 to leg segment 3 through knee and ankle pulleys in the TJ configuration. We designed four feet (\Cref{fig:collage,fig:feetcad}). The 2-segment toe (2S) design is inspired by toe phalanges of bird feet~\cite{zhang_phalangeal_2018}. We also tested a 1-segment toe (1S). 
%Toe pulley radii were selected by simulating the foot's CoP motion, by testing varying pulley radii (\Cref{fig:CoPSim}). Interesting pulley values and combination were selected based on large CoP motions \Cref{tab:leg_configurations}. 
% We selected the pulley-radii values based on simulations of the foot's CoP motion, which depend upon the torques induced by the pulley. The torque magnitudes are related to the pulley radii; hence the pulley-radii values were varied in the simulation, as seen in \Cref{fig:CoPSim}, and pulley-radii corresponding to a representative set of CoP trajectories was selected. Details related to the CoP simulation is part of a separate work and have been skipped here in the interest of space.
We selected pulley-radii values based on prior simulations we conducted of the foot's CoP motion. The foot's center of pressure moves according to instantaneous toe joint torques induced by the global tendon forces on the toe joint radii. We varied simulated values and eventually selected pulley-radii from a set of CoP trajectories \Cref{fig:CoPSim} that lead to large CoP motions. 
We further tested a ball-foot and a cylinder-foot, which are commonly used in quadruped robots \cite{katz2019mini,Grimminger_2020}. The four feet have a comparable \emph{projected} ground area (\SI{12}{\cm\squared}).

\begin{figure}
\centering
\includegraphics[width=0.42\textwidth]{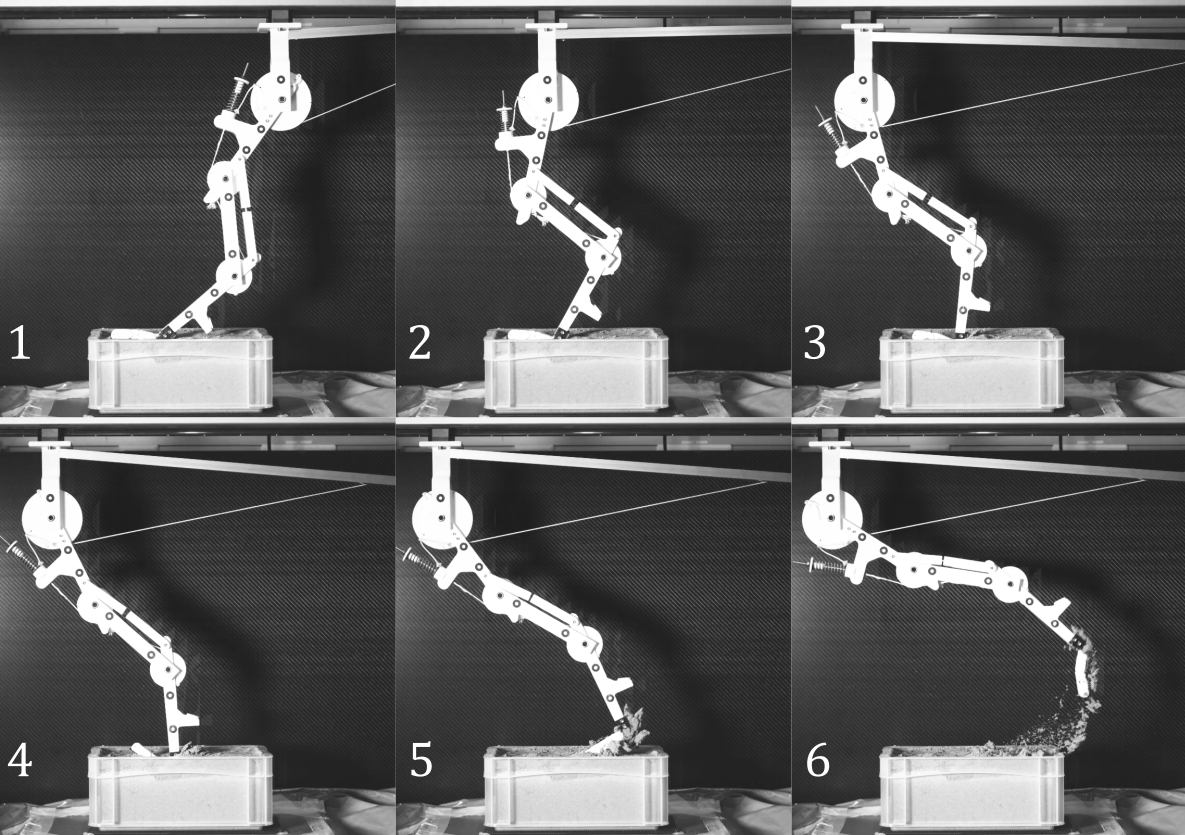}
\caption{Snapshots of one experiment on the sand, with a two-segmented toe and \SI{2}{Nm} hip torque (leg configuration LL2SC3). The leg was moved just below the slider from right to left. Here it is pushed from outside the picture with a stick, but otherwise directly by hand. Snapshot 1 shows the leg at touch-down (TD), 2 at mid-stance (MS), and 5 at toe-off (TO) due to sudden sliding. In snapshot 6, the toes have left the substrate and pulled sand off the floor due to the excessive hip torque. Typical ground reaction forces are shown in \Cref{fig:grf}.}
\label{fig:sequence}
\end{figure}

% \begin{comment}
%  \Cref{fig:CoPSim} presents the simulated CoP trajectories for a leg-length coupled leg with varying foot pulley radii. The CoP value is considered moving forward along the foot about $j3$, as shown in \Cref{fig:SegFootCoupleExample}. The foot pulley radius is varied between $r_{j3} = \SI{1}{\mm}$ and $r_{j3} = \SI{14.2}{\mm}$. The mid-stance CoP value vary between $\SI{38}{mm}$ and $\SI{3}{mm}$ for the given range of pulley radii. We selected a representative set of pulley radii values that span the range of CoP trajectories seen in \Cref{fig:CoPSim}.
% \end{comment}
%%%%%%%%%%%%%%%%%%%%%%%%%%%%%%%%%%%%%%%%%%%%%%%%%%%%%%%%%%%%%%%%%%%%%%%%%%%%%%%%
%%%%%%%%%%%%%%%%%%%%%%%%%%%%%%%%%%%%%%%%%%%%%%%%%%%%%%%%%%%%%%%%%%%
\textbf{Experimental Setup}
\label{subsec:experimentalsetup}
\begin{figure}[ht]
\centering
\includegraphics[width=0.8\columnwidth]{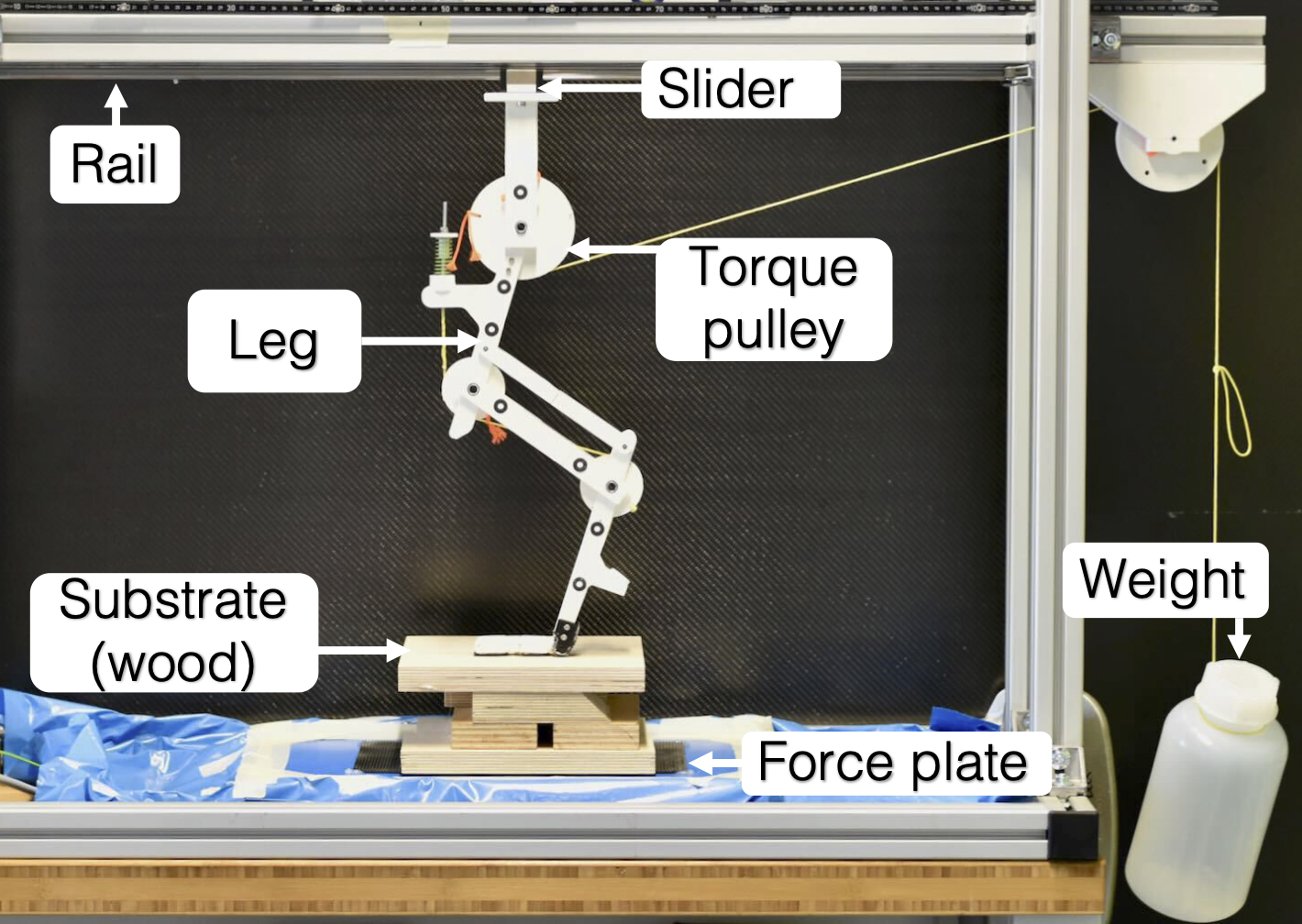}
	\caption{The experimental setup to measure maximum fore-aft (horizontal) forces and leg lengths at the onset of sliding on natural substrates. Initially the feet are manually set onto the substrate. The leg is spring-loaded. Hence, at its resting length and a starting touch-down (TD) angle of around \SI{60}{\degree} it applies the smallest ground reaction forces (GRF). The robot leg is then pushed manually from the right to the left while the hip is guided by the slider-rail (\Cref{fig:sequence}). When moving the hip forward, vertical GRFs increase until highest around mid-stance and decrease again towards toe-off (\Cref{fig:grf}). This setup simulates the single-hump ground vertical reaction and rotated-S-shaped horizontal ground reaction forces associated with running locomotion. Additional hip torque was applied through a weight on a pulley fixed to the hip. A force plate measures GRFs mounted below the substrate. We tested two rigid and two non-rigid substrates.}
\label{fig:setup}
\end{figure}
\begin{table}[b]
\caption{Leg design parameters, definitions in \Cref{fig:schematic}.}
\label{tab:leg_parameters}
\centering
\begin{tabular}{lcllc}
\textbf{Parameter} & \textbf{Value}                    &  & \textbf{Parameter} & \textbf{Value}                     \\ \cline{1-2} \cline{4-5} 
$l_{ls1}$          & \SI{160}{mm}  &  & $r_{m}$            & \SI{47}{mm}    \\
$l_{ls2}$          & \SI{160}{mm}  &  & $r_{cf}$           & \SI{25}{mm}    \\
$l_{ls3}$          & \SI{160}{mm}  &  & $r_{bf}$           & \SI{21.5}{mm}  \\
$l_{ts1}$          & \SI{39}{mm}   &  & $k_{GS}$           & \SI{2.2}{N/mm} \\
$l_{ts2}$          & \SI{39}{mm}   &  & $k_{BS}$           & \SI{5.9}{N/mm} \\
$l_{ts1s}$         & \SI{78}{mm}   &  & $k_{TJ}$           & \SI{4.0}{N/mm} \\
$r_{j1}$           & \SI{26.6}{mm} &  & $m$                & \SI{4.3}{kg}   \\ % \cline{1-2} \cline{4-5} 
\end{tabular}
\end{table}
 % table
An aluminium profile frame was rigidly mounted to a table (\Cref{fig:setup}). A rail was horizontally mounted below the top of the frame, guiding the leg in the fore-aft direction. The vertical distance between the rail and substrate surface was kept constant for all substrates by shimming the box with wooden blocks. The hip was slowly and manually pushed from side to side. The leg's foot remained standing on the substrate (\Cref{fig:sequence}). Prior, our two human experimenters practiced pushing with constant speed to minimize inertial effects. 
%The leg's hip was pushed manually---slowly---from one side to the other, with the leg's foot standing on the substrate (\Cref{fig:sequence}). 
A weight attached to a rope applied a \SI{2}{Nm} hip torque through the hip pulley. We mounted a portable force plate (9260AA, \textit{Kistler}, data acquisition system 5695B, sampling frequency \SI{250}{Hz}, BioWare software, v.5.4.8.0) under the substrate to measure the resulting ground reaction forces (GRFs). We visually recorded all experiments with a video camera (FDR-AX100, \textit{Sony}) at 4K resolution and \SI{25}{Hz} to extract optical leg markers at mid-stance and toe-off. We also recorded selected experiments with a high-speed camera (S-Motion 104M, \textit{AOS}) at \SI{1}{\kHz}.
%%%%%%%%%%%%%%%%%%%%%%%%%%%%%%%%%%%%%%%%%%%%%%%%%%%%%%%%%%%%%%%%%%%

\begin{figure}[t]
\centering
\includegraphics{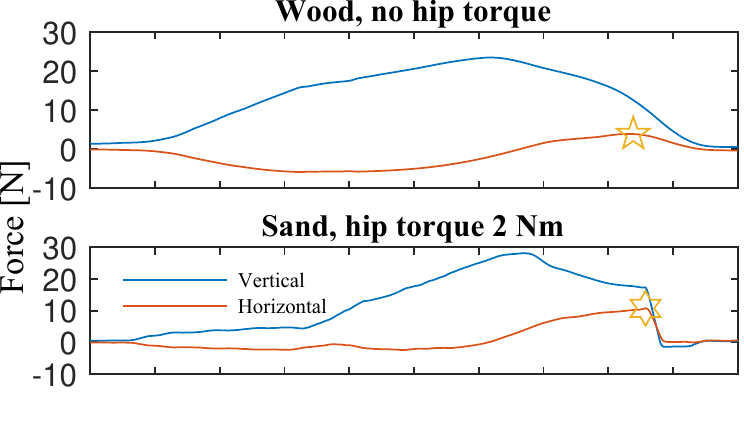}
\caption{Examples of horizontal and vertical ground reaction forces, of two experiments of the LL2SC2 configuration. In experiments without hip torque, we observed no sliding. Highest horizontal forces are indicated with a star. For the \SI{2}{Nm} hip torque experiment on sand, sliding occurred here at a horizontal peak force of \SI{10}{N}. At the mark, vertical and horizontal forces show a simultaneous, sudden drop, also corresponding to snapshot 5 in \Cref{fig:sequence}. The x-axis shows experimental time; we manually and slowly pushed the leg's hip.}
\label{fig:grf}
\end{figure}

We tested four substrates; three natural ones and a flat, \SI{12}{mm} plywood as reference (\Cref{fig:collage}). As a second hard surface, we built an uneven terrain patch by hot-glueing natural stones (\SI{30}-\SI{60}{mm}) to a plywood board. As a first non-rigid substrate, we poured small-sized quartz pebbles into a square plastic box of about \SI{15}{cm} height (\Cref{fig:sequence}). Pebbles sized between \SIrange{6}{8}{mm} and behaved like bulk solids due to their small size, low weight, and rounded edges; a foot can easily dig into them. The second non-rigid substrate was quartz sand (DIN EN 71/3) with a grain size up to \SI{2}{mm}, also often used on playgrounds. The sand was also poured into a plastic box. All stone and sand substrates were obtained from a local hardware store \cite{weco_naturstein_2022}. Two rigid and two non-rigid substrates were tested, with nine leg configurations and four substrates. We repeated each configuration with and without the hip torque on wood, pebbles and sand for five times and on stones ten times, also to reduce uncertainties from manually push of the leg. 
% The repetitions helped in reducing the
One configuration was tested 50 times, leading to 450 overall experiments. We observed no sliding in no-torque experiments (\Cref{fig:grf} top). Hence, we only report experimental results with \SI{2}{Nm} hip torque applied through the additional weight.
During soft substrate experiments sand and pebble we noticed a reduction of substrate level due to the forced spillage at TO (\Cref{fig:sequence}). Hence, we corrected vertical GRFs in post-processing via linear interpolation. 

%%%%%%%%%%%%%%%%%%%%%%%%%%%%%%%%%%%%%%%%%%%%%%%%%%%%%%%%%%%%%%%%%%%
\textbf{Ground Reaction Force Analysis}
Force plate and video data were recorded to obtain peak toe-off (TO) horizontal GRFs and leg lengths at mid-stance (MS) and TO. To compare the peak TO horizontal GRF for all legs on each substrate, we calculate the average TO horizontal GRF for all leg types on a given substrate. We denote the peak horizontal force before sliding, of a leg $l$ on a substrate $s$ as $^{s}F_{h,l}$. 
Average horizontal GRFs for all nine legs on one substrate $s$ are calculated as
\begin{equation}
    \label{eq:F_h-avg}
	^{s}F_{h,avg} \si{[N]} = \frac{1}{9} \sum_l^{} { ^{s}F_{h,l}}
\end{equation}
We define toe-off as the onset of sliding, which we detect by inspecting both GRF profiles (\Cref{fig:grf}).
The average horizontal GRF for all legs on one substrate $s$ is denoted by $^{s} F_{h,avg}$ (\Cref{eq:F_h-avg}). 
We plot horizontal GRF-values as difference from average horizontal GRFs (in \si{\%}) per substrate and for all leg configurations.
\fhlong{} (short: \fh{}) is calculated as:
\begin{equation}
\label{eq:average_hor_force}
	\fhlong{} = \fh{} = \frac{^{s} F_{h,l} \ - \ ^{s} F_{h,avg}}{ ^{s} F_{h, avg} } \cdot \SI{100}{\%} 
\end{equation}
\fh{} allows a comparison of peak horizontal forces at toe-off among the nine leg types, i.e., how well a leg can withstand slipping on a given substrate. Hence, it indicates the leg-foot's ability to accelerate better on a substrate or climb slope without sliding. 

\textbf{Leg Length Analysis}
We extracted leg lengths at mid-stance and toe-off by inspecting video recordings of experiments. We identified MS as the hip joint marker ($j_0$) vertically aligned with the toe joint 1 ($j_3$). We could directly identify the onset of sliding/TO on hard substrates wood and stone. Identifying TO on soft substrates was more challenging; we selected video frames where the foot left the soft substrate. We tagged joints $j_0$, $j_1$, $j_2$ and one marker at the bottom of leg segment 3 in snapshots in ImageJ (version 1.53k, \cite{schneider2012nih}). We extracted marker positions with a custom-written script, and processed joint angles and leg lengths by forward kinematics in Matlab.

%%%%%%%%%%%%%%%%%%%%%%%%%%%%%%%%%%%%%%%%%%%%%%%%%%%%%%%%%%%%%%%%%%%%%%%%%%%%%%%%
\section{Results}
\label{sec:results}
%%%%%%%%%%%%%%%%%%%%%%%%%%%%%%%%%%%%%%%%%%%%%%%%%%%%%%%%%%%%%%%%%%%
%%%%%%%%%%%%%%%%%%%%%%%%%%%%%%%%%%%%%%%%%%%%%%%%%%%%%%%%%%%%%%%%%%%
We conducted push-hip-torque experiments with nine leg-toe configurations on the four substrates (\Cref{tab:leg_configurations}). 
A large viable \fh{} indicates a foot that provides good traction before slipping, allowing good support for acceleration and climbing slopes. 
We separately analyse conditions that led to or prevented sinking or increased \fh{} by comparing leg performances within the class of rigid substrates (wood, stone) and soft substrates (pebbles, sand). 

\begin{figure*}
\centering
\includegraphics{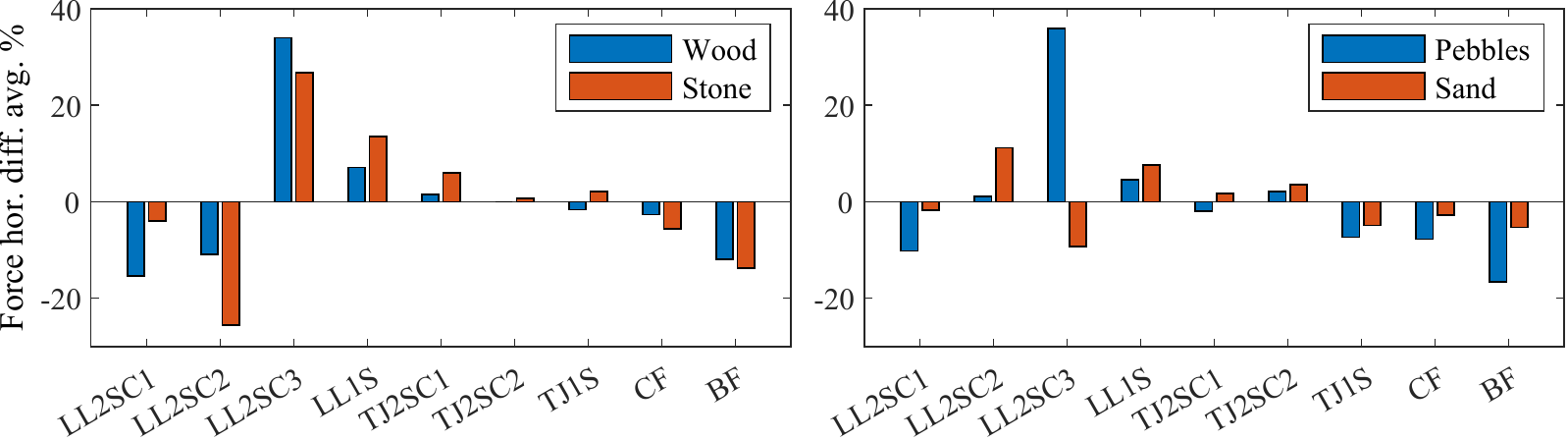}
	\caption{Deviation from average peak TO horizontal force before sliding (\fhlong{}, \Cref{eq:average_hor_force}) plotted on the y-axis, shown in percent. Figures are grouped for wood and stone (left) and pebbles and sand (right). Leg-foot configurations are listed as x-axis labels.}
\label{fig:Fhor}
\end{figure*}

Both 2-segmented LL-legs with a small j3-pulley (\Cref{tab:leg_configurations}, \SI{3.2}{mm}, LL2SC1, LL2SC2) show below-average \fh-values of \SI{-15}{\%} and \SI{-10}{\%} for wood, and \SI{-4}{\%} and \SI{-30}{\%} for stone, respectively (\Cref{fig:Fhor}, left). 
LL2SC3 with its almost three-times larger j3-pulley (\SI{9}{mm}) features high \fh{} both on wood (\SI{34}{\%}) and stone (\SI{27}{\%}). Hence, the larger j3-pulley caused a strong \fh{} improvement (\SI{49}{\%} for wood, \SI{31}{\%} for stone) for an otherwise equal j4-pulley radius between 2-segment toe configurations LL2SC1 and LL2SC3, respectively.

Leg-length coupling with a single-segment toe (LL1S) shows \fh{} values of \SI{7}{\%} (wood) and \SI{14}{\%} (stone). Hence, the single-segment toe configuration LL1S outperforms LL2SC1 and LL2SC2, whereas all members of this group feature an equal-sized j3-pulley radius.

Values for \fh{} of all toe-joint coupled legs on wood and stone (TJ2SC1, TJSC2, TJ1S) are near-zero, i.e., close to the average of all legs on these two substrates.

The cylinder- and the ball-foot performed below-average on wood, with \fh{} of  \SI{-3}{\%} and \SI{-12}{\%}. The tendency is similar on stone with \fh{} of \SI{-6}{\%} (CF) and \SI{-14}{\%} (BF). We find higher \fh-values for LL2SC2, LL2SC3, CF and BF on wood than stone. Else, stone offers higher traction for the remaining segmented leg-foot configurations of LL and TJ.

As expected, results for soft substrates pebbles and sand differ from hard substrates (\Cref{fig:Fhor} right). 
We find good \fh-values of LL2SC2 on pebbles (\SI{1}{\%}) and sand (\SI{11}{\%}). 
Like on hard surfaces, configuration LL2SC1 shows below-average \fh{} on pebbles (\SI{-10}{\%}) and on sand (\SI{-2}{\%}).
LL2SC3 shows the globally best \fh{} of \SI{36}{\%} on pebbles, yet it performed below-average on sand (\SI{-9}{\%}).
%%%%%%%%%%%%%%%%%%%%%%%%%%%%%%%%%%%%%%%%%%%%%%%%%%%%%%%%%%%%%%%%%%%
\begin{figure}[t]
\centering
\includegraphics[width=\columnwidth]{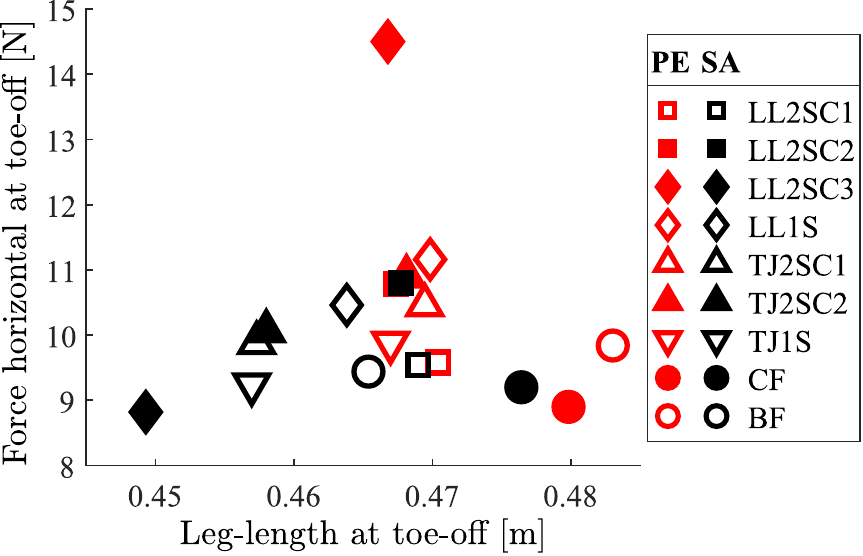}
\caption{Maximum horizontal force at the moment of sliding/TO, plotted against the TO leg length. Results are shown for the soft substrates pebble (PE) and sand (SA). Higher LL-values indicate more sinking. }
\label{fig:fh_to}
\end{figure}

With the exception of LL2SC3, leg configurations show overall better \fh{} on sand, compared to pebble. We conclude that the sand substrate provided, in most cases, higher traction than the loose-pebble substrate. 

Arguably, the efficiency of legged locomotion on soft substrate also depends on the foot-/toe-ability to reduce sinking. We expect that sinking will negatively affect locomotion efficiency, speed, and stability. The hip position in our setup is slider-guided on a constant-height, horizontal rail (\Cref{fig:setup}). Hence, the amount of sinking is related to leg lengths for similar TO leg angles (small-angle approximation), and larger leg lengths imply more sinking. \Cref{fig:fh_to} plots horizontal leg forces against leg lengths at toe-off, for pebble (red markers) and sand (black markers). 

The LL2SC3 configuration shows a \SI{15}{\newton} high horizontal force and a small toe-off leg length (little sinking), compared to all non-segmented feet on pebbles. LL2SC3 sinks on pebbles comparably to other segmented legs, with a LL of \SI{46}{\cm} (\Cref{fig:fh_to}).
On the sand, LL2SC3 has a low toe-off LL of about \SI{45}{\cm} and a low horizontal toe-off force around \SI{9}{\newton}. Both remaining 2-segmented LL coupling legs (LL2SC1, LL2SC2) show little difference between pebble and sand substrates (\Cref{fig:fh_to}).
There are differences in the behaviours of other segment-toes LL1S, TJ1S, TJ2SC1 and TJ2SC2, all of which increase horizontal forces by sinking more into pebbles than sand. Cylinder- and ball-feet sank more into pebbles than sand, between \SIrange{0.5}{2}{\cm}.
Overall, we find that more sinking did not correspond to gains in horizontal force (\Cref{fig:fh_to}).

%%%%%%%%%%%%%%%%%%%%%%%%%%%%%%%%%%%%%%%%%%%%%%%%%%%%%%%%%%%%%%%%%%%
\section{Discussion}\label{sec:discussion}
We were aiming to identify foot and leg parameters that enable high horizontal GRF before slipping. We varied number of foot segments (1, 2), leg length coupling type (LL, JL), foot type (BF, CF) and pulley-radii. The 2-segment foot with leg-length coupling (LL2SC3) and large j3-foot-pulley radius stands out. While performing below-average on sand, this configuration outperformed all legs on wood, stone, and pebbles. 

We observed that ball- and cam-feet sank into sand/pebble substrates more as they rotate during leg vaulting; these feet dug into the ground. In comparison, segmented feet largely decouple leg rotation and ground support. We observed how much foot pulley selection drives the foot segment behaviour through foot joint torques (LL2SC3). We expect that we can achieve a similar effect by changing pulley-radii (i.e., using nonlinear cam surfaces), toe segment lengths like observed in birds, or supplying or removing additional torque to the foot segments through in-parallel, active actuation. Further, our experiments document horizontal force margins on various legs. These can serve as the basis for active control of foot segments in robot legs during locomotion over natural terrains. 
Some of our questions about locomotion remain unanswered. For example, our setup was not ideal to measure friction cones; the hip height of the legs was fixed, which limited the range of leg length and, thereby, vertical force. Hence, a setup for friction cone measurement should supply a variable hip height, likely in the form of a legged, hopping or running robot with instrumentation for horizontal and vertical GRF. 
As an alternative to experiments in hardware, reduced-order or complex models of natural substrates can be computed. Due to many variables and nonlinear and complex interaction characteristics between substrate and robots, these methods require long run times and hardware validation \cite{falkingham_birth_2014,aguilar_robophysical_2015}. We see our work as an additional hardware data set in line with previous experiments on soft and fluidic substrates \cite{liTerradynamicsLeggedLocomotion2013,gongKinematicGaitSynthesis2016,sharpeLocomotorBenefitsBeing2014,goldmanMarchSandbots2009,liMultifunctionalFootUse2012,qianPrinciplesAppendageDesign2015,liSensitiveDependenceMotion2009}. Especially for the case of legged locomotion on natural substrates, we believe that hardware experiments and computer simulations should consequitively inform each other.

\section{Conclusion}\label{sec:conclusion}
We report that two-segmented, mechanically adaptive feet show increased peak horizontal forces between configurations and on soft and hard substrates before starting to slip, with increases of viable horizontal force up to \SI{49}{\%} caused by selecting appropriate toe joint pulley radii associated with higher toe joint torques. Segmented feet reduced sinking on soft substrates compared to ball- or cylinder-feet, by almost \SI{2}{cm} or \SI{5}{\%} mid stance leg length. Our proposed, mechanically adaptive, multi-segmented feet provide a wide range of centre of pressure points that are well suited for bipedal robots, but also for quadruped robots locomoting in soft terrains. We expect that an actuated, multi-segment toe, e.g. with an in-parallel tendon connected to a motor, will provide the full range of peak TO horizontal forces reported here. The proposed two-segmented feet are lightweight, robust, and mechanically simple to design and implement. We plan to integrate them into our two- and four-legged robots to locomote in natural terrains.

%%%%%%%%%%%%%%%%%%%%%%%%%%%%%%%%%%%%%%%%%%%%%%%%%%%%%%%%%%%%%%%%%%%%%%%%%%%%%%%%
\section*{Acknowledgment and supplementary files}
This work is funded by the Deutsche Forschungsgemeinschaft (DFG, German Research Foundation) 449427815, 449912641 and the Max Planck Society. The authors thank the International Max Planck Research School for Intelligent Systems (IMPRS-IS) for supporting An Mo, Bernadett Kiss, and Emre Cemal G\"onen. An Mo is supported by the China Scholarship Council. We thank Patrick Frank for his support in designing and assembling the experimental setup.
% A brief supplementary section is available at the end of this linked \href{www.link-to-keeper-url.de}{file}.
A high resolution version of the submitted video file is available at \href{https://youtu.be/tJ7v81YDGRE}{https://youtu.be/tJ7v81YDGRE}.
CAD files to replicate experiments are available through \href{https://doi.org/10.17617/3.A6VVVU}{https://doi.org/10.17617/3.A6VVVU} (CC 4.0 BY-NC).
%%%%%%%%%%%%%%%%%%%%%%%%%%%%%%%%%%%%%%%%%%%%%%%%%%%%%%%%%%%%%%%%%%%%%%%%%%%%%%%%
% \bibliographystyle{IEEEtran.bst}
% \newpage
\bibliographystyle{IEEEtran}
\bibliography{main.bib}

% \input{Tables/tab_leg_length_change_pe_sa}
% %%%%%%%%%%%%%%%%%%%%%%%%%%%%%%%%%%%%%%%%%%%%%%%%%%%%%%%%%%%%%%%%%%%%%%%%%%%%%%%%
% \input{Tables/tab_appendix_average_horizontal_force}

\addtolength{\textheight}{-12cm}   % This command serves to balance the column lengths
\end{document}